\documentclass{article}
\usepackage{spconf,amsmath,graphicx}

\usepackage{hyperref}       
\usepackage{url}   
\usepackage{comment}
\usepackage{color}
\usepackage{kotex}
\usepackage{adjustbox}

\usepackage{booktabs}       

\usepackage{multicol}
\usepackage{multirow}
\usepackage{amsmath}
\usepackage{mathtools}
\usepackage{amssymb}
\usepackage{comment}

\usepackage{caption}
\usepackage{subcaption}
\usepackage{xcolor}

\usepackage{url}

\usepackage[absolute,showboxes]{textpos}

\TPMargin{5pt}

\newcommand{\copyrightline}{
    \begin{textblock}{0.84}(0.08,0.93)    
         \noindent
         \footnotesize
         \copyright 2021 IEEE. Personal use of this material is permitted. Permission from IEEE must be obtained for all other uses, in any current or future media, including reprinting/republishing this material for advertising or promotional purposes, creating new collective works, for resale or redistribution to servers or lists, or reuse of any copyrighted component of this work in other works.
    \end{textblock}
}

\title{Prototype-based Personalized Pruning}
\name{Jangho Kim $^{*,2}$
\thanks{\scriptsize{${}^{*}$Author completed the research in part during an internship at Qualcomm Technologies, Inc. ${}^{\dagger}$Qualcomm AI Research is an initiative of Qualcomm Technologies, Inc.}}
\qquad Simyung Chang $^{1}$ \qquad Sungrack Yun $^{1}$ \qquad Nojun Kwak $^{2}$}

\address{$^{1}$Qualcomm AI Research${}^{\dagger}$, Qualcomm Korea YH\\
$^{2}$ Seoul National University \\
{\small\texttt {kjh91@snu.ac.kr, \{simychan, sungrack\}@qti.qualcomm.com, nojunk@snu.ac.kr} }
}

\begin{document}
%
\maketitle
\copyrightline
\begin{abstract}
Nowadays, as edge devices such as smartphones become prevalent, there are increasing demands for personalized services. However, traditional personalization methods are not suitable for edge devices because retraining or finetuning is needed with limited personal data. Also, a full model might be too heavy for edge devices with limited resources. Unfortunately, model compression methods which can handle the model complexity issue also require the retraining phase. These multiple training phases generally need huge computational cost during on-device learning which can be a burden to edge devices. 
In this work, we propose a dynamic personalization method called \textit{prototype-based personalized pruning} (PPP). PPP considers both ends of personalization and model efficiency. After training a network, PPP can easily prune the network with a prototype representing the characteristics of personal data and it performs well without retraining or finetuning. We verify the usefulness of PPP on a couple of tasks in computer vision and Keyword spotting.

\end{abstract}
\section{Introduction}
\label{sec:intro}

Personalization is an important topic in computer vision and signal processing with many applications such as recommender systems, smart assistance, speaker verification and keyword spotting \cite{xue2014singular,he2019device,sarikaya2017technology}.
Personalization of a global model may have conflicting objectives in terms of generalization and personalization. In other words, a global model should perform well not only on general data but also on personal data. In order to train a global model to perform well in many circumstances, the training data should cover all the cases for generalization. However, this assumption departs from the real world scenario which can not cover every cases. One of naive approaches to personalize a deep neural network is finetuning the global model with limited personal data. However, it is not practical to store a global model on memory-restricted edge devices.

Recently, the deep neural network (DNN) models have been widely adopted in various domains such as computer vision, audio processing, text analysis, and gene data expression.
Deploying a DNN requires high computational resources. Meanwhile, a trend of deploying a DNN have moved from a desktop computer into edge devices and smartphones. However, deploying a DNN on edge devices is a challenging problem because of its limited memory and computational resources. 
To resolve this issue, many model compression methods have been widely studied such as knowledge distillation \cite{kim2018paraphrasing,hinton2015distilling,kim2019feature}, model quantization \cite{courbariaux2016binarized,rastegari2016xnor} and model pruning \cite{han2015deep,kim2020position}. Most of model compression methods normally need an additional training or retraining phase. Therefore, an additional computational cost is unavoidable for a model considering both personalization and model efficiency. 

\begin{figure}[t]
  \centering
  \includegraphics[width = 1\linewidth]{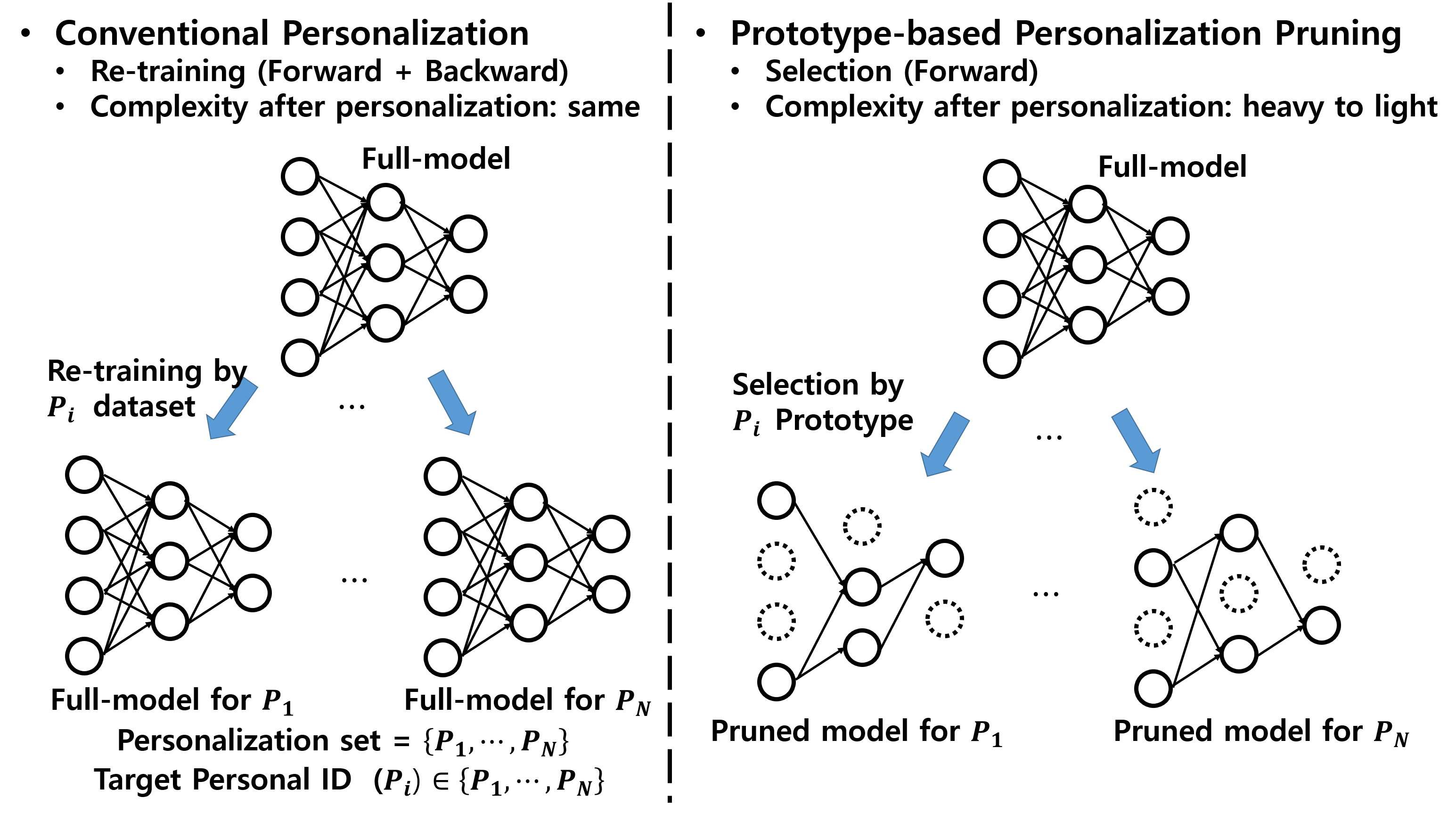}\\
  \caption{The differences between conventional personalization and Prototype-based Personalized Pruning (PPP).  }
    \label{fig:different}
\end{figure}

\begin{figure*}[h]
\begin{minipage}[b]{0.6\linewidth}
\includegraphics[width=0.9\textwidth]{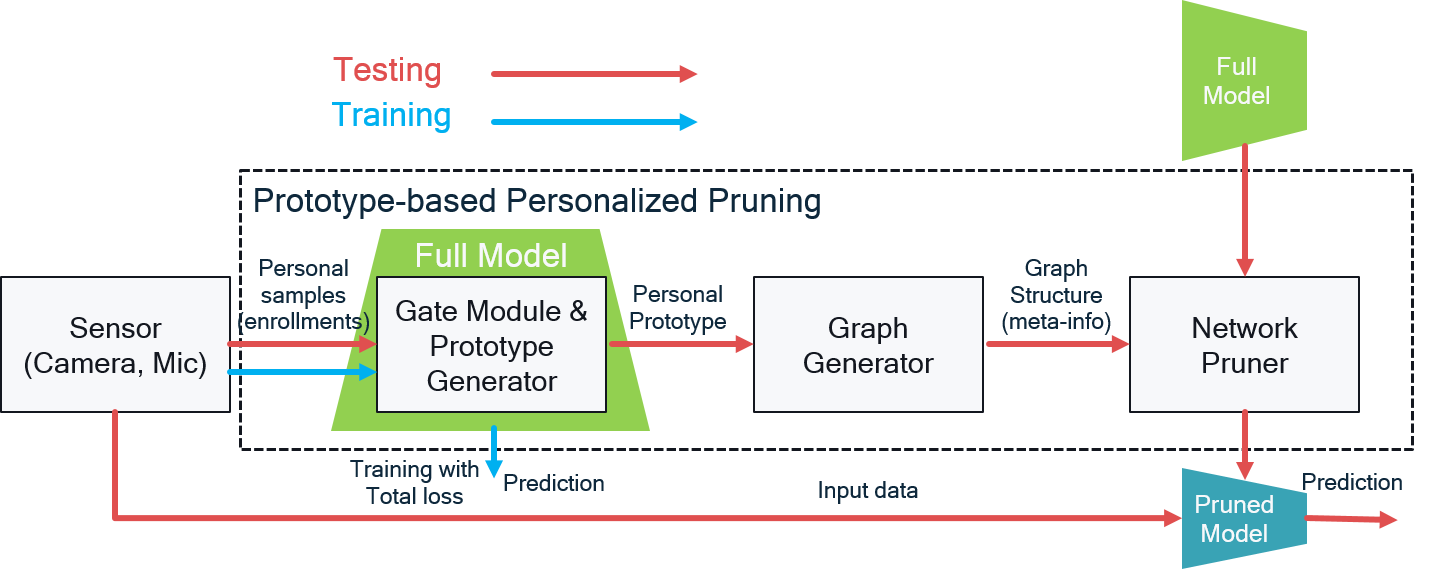}
\caption{The overall process of PPP.}
\label{overallprocess}
\end{minipage}
\begin{minipage}[b]{0.38\linewidth}
\centering
\includegraphics[width=0.6\textwidth]{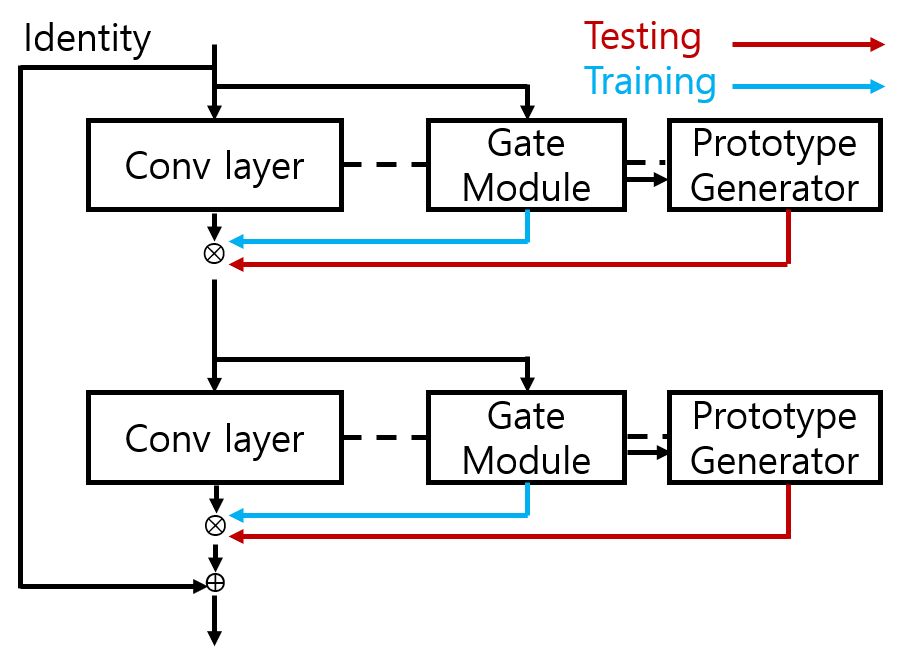}
\caption{Residual block containing gate modules and prototype generators.}
\label{gatemodule}
\end{minipage}
\end{figure*}

Our goal is to train the global model that can be easily pruned with personal data and perform well after pruning without any finetuning. To achieve this goal, we propose a novel method for personalization using a dynamic path network. The proposed method is named as prototype-based personalized pruning (PPP) inspired by prototypical network \cite{snell2017prototypical} which learns a metric space using prototype representation. PPP considers the model complexity as well as personalization. After training, unlike dynamic path network \cite{veit2018convolutional}, PPP selects a subset graph with a prototype module using the prototype driven from limited personalized data. For this reason, the model structure of ppp is dynamically customized to enhance personalization and reduce the model complexity. The differences between conventional personalization and PPP is depicted in Fig.\ref{fig:different}. More details are explained in Sec. \ref{sec:Method}.

\section{Related work}
\label{sec:relatedwork}
\noindent \textbf{Dynamic path network} The computational graph in a general DNN is fixed. On the other hand, the works in \cite{lee2020urnet,lin2017runtime,veit2018convolutional} propose dynamic path networks. Although a full network is needed, they select the forwarding graph (set of path) at inference time.  Utilizing this may help the network to change its complexity at inference time by reducing the forward graph. However, selecting the forward graph of these methods depends on the incoming input. Therefore, they have to store a full model, which is not practical in terms of deploying the complex full model on a resource constraint edge device. 

\noindent \textbf{Prototypical network} 
Prototypical network \cite{snell2017prototypical} focuses on learning an embedding space where data can be classified by computing the distance to a prototype representation of each class. This prototypical framework is also used in speaker recognition with metric learning \cite{wang2019centroid} and few-shot learning domain \cite{jadon2020overview}. 

In this work, we adopt the above two frameworks. We make a prototype per personality (each identity or group of personal data) and train a dynamic path network for personalized pruning. After training, PPP provides the personal model which is a pruned version of the full model on the fly without finetuning with given personal data.

\section{Proposed method}
\label{sec:Method}
Our PPP method is based on the AIG \cite{veit2018convolutional} which is a dynamic path network and the prototypical network \cite{snell2017prototypical}. Similar to AIG, in the training phase, PPP trains a gate module to select proper channels for pruning. At the same time, we also consider personalization. We define the prototype of personalization using the output of the gate module which is the binary embedding pattern for each convolution module indicating whether to prune each channel or not. To consider the performance of after-pruning with prototype at the training phase, we regularize the output of the gate module by making it similar to the prototype. As a result, after training, we can easily prune the global model (full model) by pruning with prototype derived from the given restricted personal data using graph generator and network pruner. PPP does not need the global model after training because the channel selection is based on the prototype representing the personal data and is not based on an incoming input as in traditional dynamic path networks. The overall process is depicted in Fig. \ref{overallprocess}.

\subsection{Gate module}
For the sake of explanation, we set a base network as a ResNet \cite{he2016deep} but it is not restricted to a residual network architecture.
Unlike AIG \cite{veit2018convolutional}, our gate module controls each channel of a convolution module not a whole residual block (See Fig. \ref{gatemodule}). 
Let $g$ be a gate per each conv layer.
${g}: {x} \in \mathbb{R}^{n \times w \times h} \rightarrow {z} \in \mathbb{R}^m$ 
where $x$ is an input to a gate module and $z$ is an embedding vector in the gating embedding space. $n$ and $m$ are the number of input and output channels of the conv layer related to the gate module. We use spatial global average pooling to the input $x$.  The embedding vector $z$ learns whether or not to use an output channel with a probability which is a random variable with a range of $[0,1]$. We use Gumble Softmax and straight through estimator for training gate modules \cite{jang2016categorical}. In other words, in the forward pass, we use argmax with a probability $p$ for making hard attention ($z \in \{0,1\}$) but the gradient is calculated by 
the Gumble-Max trick \cite{veit2018convolutional}. Except the output dimension, other structures of gate module is the same as in AIG \cite{veit2018convolutional}.

\subsection{Training }
\label{subsec:training}
Let the training dataset be $D = \{(x_i, t_i, p_i)_{i=1}^N$\}, $t_i\in\{1,\cdots,\mathcal{K}\}$, $p_i \in \{1, \cdots, \mathcal{P} \}$, where $\mathcal{K}$ and $\mathcal{P}$ are the number of classes and the number of identities (individual clients), respectively\footnote{The class and the identity might be identical for certain tasks such as face recognition but it can be different for other tasks such as keyword spotting. For generality, we separate these two.}. For brevity, we use the notation $(x^p_i$,$t^p_i)$ to denote the $i$-th data pair corresponding to the $p$-th identity. We can consider a conv layer as ${f}: {x} \in \mathbb{R}^{n\times w \times h} \rightarrow {y} \in \mathbb{R}^{m\times w \times h}$ where $x$ is an input activation and $y$ is the corresponding output of the conv layer. The gate module $g$ receives the same input $x$ and it becomes
\begin{equation}
\label{eq:1}
y^p_{i,l}=f(x^p_{i,l};\theta_{f,l}), \qquad z^p_{i,l}=g(x^p_{i,l};\theta_{g,l}). 
\end{equation}
Where $\theta_{f,l}$ and $\theta_{g,l}$ are the parameters for the $l$-th conv layer and gate module respectively.
We can calculate the output feature map $y$ and the embedding vector $z$ using the incoming input.  
To train the gate module for channel selection, we calculate the pruned channel by multiplying each channel of the output $y$ with the corresponding element of the embedding vector $z$ for every conv layer. 

So far, the training strategy is similar to previous dynamic path networks in terms of channel selection based on an incoming input. Here, we introduce the definition of `Prototype' representing the personal identity. In each minibatch, let the number of samples with $p$-th identity be $n_p$. Then, we can calculate the mean of embedding vectors among all $z^p$'s. We define $\Bar{z}^p$ as this mean in a minibatch 
which represents the personality of $p$.
\begin{equation}
\label{eq:2}
\Bar{z}^p= \frac{1}{n_p}\sum_{i=1}^{n_p} z^p_i 
\end{equation}
We calculate the set of $\{\Bar{z}^p_l\}$ for all conv layer at the prototype generator and this set is used as a prototype of $p$. Here, the subscript $l$ denotes the $l$-th layer. This prototype is not a discrete vector because of the mean operation. At test time which requires a pruned model, it needs to be a discrete vector. We can make a prototype as a discrete vector using element-wise step function $\mathcal{S}:\mathbb{R}^m \rightarrow \mathbb{R}^m$ with a threshold value $\tau$.
\begin{equation*}
\forall i \in \{1,\cdots, m\}, \quad \mathcal{S}(x_i)_i = \begin{cases}
1 &\text{$x_i$ $\geq$ $\tau$}\\
0 &\text{Otherwise}
\end{cases}
\end{equation*}
Let $\mathcal{C}$ be the set of conv layers in the network and $\mathcal{B}$ is a mini batch containing a subset of identities. Then, we introduce the regularization loss for personlization.
\begin{equation}
\label{eq:3}
\mathcal{L}_{prototype}= \frac{1}{|\mathcal{C}|}\sum_{l\in \mathcal{C}}\frac{1}{|\mathcal{B}|}\sum_{p=1}^\mathcal{P}\sum_{i=1}^{n_p} \lVert z^p_{i,l}-\mathcal{S}(\Bar{z}^p_l) \rVert^2_2, 
\end{equation}
This loss regularizes each input with the same identity to output a similar gate pattern not deviating much from the corresponding prototype $\mathcal{S}(\Bar{z}^p_c)$. This can make PPP possible to prune the network based on an identity-specific prototype $\mathcal{S}(\Bar{z}^p_l)$, not by a specific input. 
In this regularization, we do not use the distance metric between prototypes because there is no reason to broaden the distance among different prototypes like \cite{snell2017prototypical}. 

We use a soft constraint for the utilization rate of a network by introducing a target loss. We count the number of alive channel (filter) in each conv layer and regularize this rate with a specific target $T$: 
\begin{equation}
\label{eq:4}
\mathcal{L}_{target}= \left(T-\frac{1}{|\mathcal{C}|}\sum_{l\in \mathcal{C}}\frac{1}{|\mathcal{B}|\times m_l}\sum_{p=1}^\mathcal{P}\sum_{i=1}^{n_p} \lVert z^p_{i,l}\rVert_1\right)^2,
\end{equation}
where $m_l$ is the number of channel in the $l$-th layer. This target loss provides an easy way to adjust network complexity.

Using the standard cross-entropy loss, $\mathcal{L}_{task}$, the overall training loss is
\begin{equation}
\label{eq:5}
\mathcal{L}_{total}=  \mathcal{L}_{task} + \alpha\mathcal{L}_{prototype} + \beta\mathcal{L}_{target}, 
\end{equation}
where $\alpha$ and $\beta$ are the hyper-parameters for balancing the losses. The training framework of PPP resembles the dynamic path network because the gate module is trained with independent input samples with the task loss. However, at the same time, the output of the gate module is regularized with the representative prototype of a specific identity by $\mathcal{L}_{prototype}$. In doing so, PPP can provide a personalized inference path by selecting a similar path within the same identity.   

\subsection{Testing }
Data-dependent path has a drawback that it always needs the full network. It selects the forward graph depending on the input. On the other hand, after the training, PPP does not use the data-dependent path. Instead, PPP makes a forward graph depending on the prototype of given personal data. Therefore, unlike traditional personalization schemes such as finetuning the global model with personal data, PPP can make use of a light-weight personal model on the fly.

At test time, with a few given personal data, we can calculate the prototype using (\ref{eq:2}). With the calculated prototype, one can obtain a personalized pruned model by eliminating the conv layer filters with the given binary pattern for each specific identity using the graph generator and network pruner, without additional training as shown in Fig.~\ref{overallprocess}.

\section{Experiments}
\label{sec:experiment}
To verify the proposed PPP, we conduct experiments on two tasks: image classification and keyword spotting (KWS). This shows PPP can handle both images and signals as input data. In this work, we choose a residual network as a baseline network. We compare our PPP with the conventional dynamic path network and the vanilla network in terms of accuracy and utilization rate which is the rate of alive parameters in conv layers of network (\# of alive parameters / \# of the total parameters). 

\subsection{Experimental Setup}
\label{subsec:setup}
In all experiments, we use the same hyper-parameters: $\alpha = \beta=10$, $\tau = 0.7$ and $T=0.6$. In the testing phase of PPP, we use the first test minibatch for the enrollment of personal data to make a prototype per identity. Then, we measure the test accuracy based on the customized personal model.

\noindent \textbf{Image Classification}
In the image classification task, 
we use CIFAR-10/100 \cite{krizhevsky2009learning} datasets with the widely-used ResNet-56 \cite{he2016deep} as a baseline network. Each CIFAR dataset contains 50,000 training data and 10,000 testing data. CIFAR-10 and 100 have 10 and 100 classes, respectively. Since there is no personal identity label in CIFAR datasets, we assume that the task label (classification label) also represents the personal identity\footnote{It corresponds to $t=p$ and $ \mathcal{K}=\mathcal{P}$ in Sec \ref{subsec:training}.}. PPP starts from the pretrained model. The initial learning rates of 0.1 and 0.01 are used for the gate module and the network, respectively. Then, we follow the training details of CIFAR experiments in \cite{he2016deep}. 

\noindent \textbf{Keyword Spotting}
For keyword spotting, we use Qualcomm Keyword Speech Dataset\footnote{\url{https://developer.qualcomm.com/project/keyword-speech-dataset}} \cite{kim2019query} and ResNet-26 \cite{tang2018deep} for the baseline network. Qualcomm keyword dataset contains 4,270 utterances of four English keywords spoken by  42-50 people. The four keywords are `Hey Android', `Hey Snapdragon', `Hi Galaxy' and `Hi Lumina'. In this experiment, unlike CIFAR datasets, the identity labels exist ($t\neq p$ and $ \mathcal{K} \neq \mathcal{P}$). We set the number of personal identity as 42 ($\mathcal{P}=42$) to fully utilize the four classes ($\mathcal{K}=4$). 
Each \textsc{wav} file is recorded with the sampling rate of 16 kHz, mono channel in 16 bits. The dataset is split into 80\% train and 20\% test data. We train the model from scratch with 100 epochs with the initial learning rate of 0.1 and decay it with a factor of 0.1 at 300 and 500 iteration. All other training details and experimental settings are identical to that of ResNet-26 in \cite{tang2018deep}.

\begin{table}

\centering
  \caption{
  Test accuracy and utilization rate
  }
\label{CIFAR_EX}
\begin{adjustbox}{width=1\linewidth}
   \begin{tabular}{  l c c c}
         \toprule
          \multicolumn{4}{c}{ResNet-56 on CIFAR-10/100 dataset.} \\
          \midrule
              {Method}  & Type & {Accuracy (\%)} & {Utilization rate (\%)}  \\
              &  & {C-10 / C-100} & {C-10 / C-100}  \\
         \midrule
   
             ResNet-56-Vanilla \cite{he2016deep} & N/A   & 92.9 / 67.9  & 100.0 / 100.0      \\
                      ResNet-56-AIG \cite{veit2018convolutional} &  Single & 92.1 / 67.8  & 60.3 / 74.0     \\
                      
                                                                    
                      ResNet-56-PPP &Single   & 92.7 / 66.7  & 40.2 / 52.0  \\
                                              \textbf{ResNet-56-PPP} &\textbf{Prototype} & 94.4 / 68.7  &   37.6 / 52.4  \\

        \bottomrule
       \toprule
         \multicolumn{4}{c}{ResNet-26 on Qualcomm Keyword Speech Dataset} \\
          \midrule
              {Method}  & Type& {Accuracy (\%)} & {Utilization rate (\%)}  \\
         \midrule
 
             ResNet-26-Vanilla \cite{tang2018deep} & N/A & 99.7  & 100.0      \\
                      
                      ResNet-26-PPP NoReg &Single   & 98.9  & 49.3     \\
                                              ResNet-26-PPP NoReg &Prototype & 53.7  & 32.2     \\
                                                                    
                      ResNet-26-PPP& Single   & 99.4  & 37.8     \\
                                              \textbf{ResNet-26-PPP} &\textbf{Prototype} & 99.4  & 35.4     \\

        \bottomrule
   \end{tabular}   
\end{adjustbox}
\end{table}

\subsection{Experimental Results}
\label{subsec:Results}
We calculate the mean accuracy and the utilization rate for each personlized model in Table \ref{CIFAR_EX}. In the table, the \textit{vanilla} means the original network trained with cross-entropy. `Type' in the table indicates the way of selecting the forward graph. `Single' implies selecting graph with independent incoming input as AIG \cite{veit2018convolutional} does. On the other hand, `Prototype', used in the proposed method, selects channels with the prototype so as to prune the model according to the personal identity. Note that in a real application, conventional dynamic networks such as AIG always need a full model per client so the utilization rate is only valid at inference time. On the other hand, in our PPP, after providing the first minibatch to the full model to determine the identity, we can obtain a pruned model based on the estimated identity and abolish the full model.

\noindent \textbf{Image Classification}
As shown in Table \ref{CIFAR_EX}, 
CIFAR-10 and 100 show similar trends. As expected, the vanilla model with full utilization outperforms both PPP (single) and AIG both of which select a different forward path for each independent input. Interestingly, PPP (Prototype), the proposed method, beats the vanilla and AIG with a large margin using less memory (low utilization rate). Compared to AIG, PPP is 2.3\% and 0.9\% more accurate on CIFAR-10 and -100, respectively. Also, PPP uses 20\% less parameters than AIG.   

\begin{figure}[t]
\centering
\includegraphics[width=0.85\linewidth]{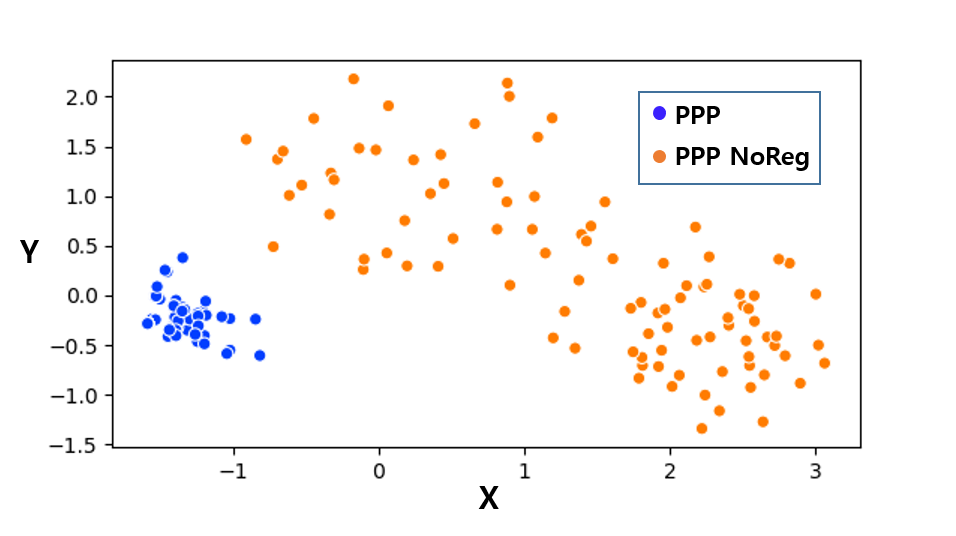}
\vskip -0.1in
\caption{Embedding vectors visualized by PCA. We plot the output of gate module attached 15-th conv layer with a same identity. Blue is from PPP and orange is from PPP NoReg. }
\vskip -0.1in
\label{fig:PCA}
\end{figure}

\noindent \textbf{Keyword Spotting}
To confirm the importance of the regularization term $\mathcal{L}_{prototype}$ for personalized pruning, we conduct an ablation study of it. 
`PPP NoReg' represents training with total loss without $\mathcal{L}_{prototype}$. PPP shows very close performance regardless of `Type'. On the other hand, in PPP NoReg cases, although the `Single' shows reasonable performance in both accuracy and utilization rate, the `Prototype' suffers from severe accuracy degradation, showing 45\% less accuracy than PPP. This is because without $\mathcal{L}_{prototype}$, the output vector of the gate module cannot be merged near the prototype (See Fig. \ref{fig:PCA}) and the pruned model using the prototype cannot retain performance.

\section{Conclusion}
\label{sec:conclusion}
In this paper, we propose a novel framework which is able to prune a full network keeping the personality, named prototype-based personalized pruning (PPP). We verify the effectiveness of PPP in image classification and keyword spotting tasks in terms of model complexity and performance. Unlike conventional dynamic path networks which need a full model, PPP can prune the network based on the identity by using the prototype. This helps considering both personalization and model complexity. We believe that PPP will help further researches in model personalization and pruning.

\vfill\pagebreak

\clearpage

\bibliographystyle{IEEEbib}
\bibliography{refs}

\end{document}